# Solving Sparse Finite Element Problems on Neuromorphic Hardware

**Authors:** Bradley H. Theilman[1]*, James B. Aimone[1]*

[1]Neural Exploration and Research Laboratory, Sandia National Laboratories; Albuquerque, NM, USA

*Corresponding author. Email: bhtheil@sandia.gov; jbaimon@sandia.gov

We demonstrate that scalable neuromorphic hardware can implement the finite element method, which is a critical numerical method for engineering and scientific discovery. Our approach maps the sparse interactions between neighboring finite elements to small populations of neurons that dynamically update according to the governing physics of a desired problem description. We show that for the Poisson equation, which describes many physical systems such as gravitational and electrostatic fields, this cortical-inspired neural circuit can achieve comparable levels of numerical accuracy and scaling while enabling the use of inherently parallel and energy-efficient neuromorphic hardware. We demonstrate that this approach can be used on the Intel Loihi 2 platform and illustrate how this approach can be extended to nontrivial mesh geometries and dynamics.

Neuromorphic computing (NMC) seeks to emulate architectural and algorithmic features of the brain to achieve lower-power, higher-capability microelectronics, as well as computing platforms. Despite this tremendous potential, the widespread impact of neuromorphic computing has been limited by the difficulty in identifying applications that show clear performance advantages [1], particularly in the realm of scientific computing [2]. There are many approaches to brain inspiration, and these provide distinct challenges to achieving the necessary performance requirements to impact real-world computing applications [3]. While analog processing-in-memory hardware is still limited by scaling obstacles, it was recently demonstrated that memristive crossbars can reach arbitrary precision for numerical dense linear algebra tasks [4]. In contrast, modern second-generation neuromorphic systems that leverage digital complementary metal–oxide–semiconductor (CMOS) architectures, such as Loihi 2 and SpiNNaker 2, are now approaching over 1 billion neurons, a similar scale to many small mammalian and avian brains [5], suggesting that spiking neuromorphic systems could potentially reach scales necessary to impact scientific simulations that currently require thousands of conventional processors.

We have previously observed a neuromorphic advantage in modeling discrete-time Markov chain random walks, specifically in applications which require integration over many Monte Carlo samples [6]. While this approach is promising for application domains whose problems can be formulated as stochastic differential equations, there remains a need to identify an efficient neuromorphic algorithm for the much broader class of partial differential equations (PDEs) solved through deterministic numerical solvers.

Here, we introduce a spiking neuromorphic algorithm that solves the sparse linear systems arising from the finite element method (FEM) for solving PDEs, and demonstrate our algorithm on Intel's Loihi 2, a modern NMC platform. Finite element methods are arguably the most predominant numerical method used in modern high-performance computing applications. We show not only that FEM can be formulated as a brain-inspired neural algorithm, but that this algorithm is scalable and highly aligned with many of the brain-inspired design choices for modern NMC systems. Importantly, unlike previous NMC algorithms, the FEM approach we show is mathematically equivalent to the standard formulations used in the numerical computing community—the only difference is how the sparse linear system is solved, a difference that is largely invisible to the user. Combined, these results demonstrate that the significant low-power advantages of NMC can be leveraged to perform energy-efficient numerical computing simulations.

**Results**

**Finite Elements**

PDEs describe physical phenomena on a domain of definition, such as understanding electrostatic forces between molecules, turbulent water flow through a turbine, or how wireless signals propagate through a building. To solve these equations, FEMs work by first discretizing the domain with a mesh (a collection of smaller geometric elements that, when combined, fully represent the domain), then approximating the solution using a linear combination of basis functions supported on mesh elements. Mathematically, applying the *weak form* of the PDE to this finite-dimensional approximation yields a sparse linear system (Ax=b) for the coefficients of this linear combination (Figure 1A). The solution of this sparse linear system yields the best linear combination of basis functions to approximate the solution of the original PDE.

To achieve high-fidelity solutions, the FEM mesh must discretize the domain at a high resolution with many elements. The associated linear systems are very large: modern FEM problems can contain many millions or even billions of variables. Importantly, because the physical interactions described by PDEs are local and the basis functions are usually compactly supported, elements in the mesh only directly interact with their nearest neighbors. This is the source of the sparsity in the associated linear system.

These large linear systems are traditionally solved with two classes of algorithms: iterative and direct solvers. Direct solvers, such as lower–upper (LU) factorization, modify the linear system in place to directly compute the solution. Iterative solvers, in contrast, compute a sequence of approximations that converge to the true solution over time. Each approach has distinct advantages, but iterative solvers tend to be the choice for truly large-scale problems [7].

Our neuromorphic FEM solver is analogous to traditional iterative solvers in that we construct a spiking neural network with specific dynamics that converge to the solution of the linear system over time. Our neuromorphic formulation, which does not require learning or training, is intrinsically parallel and benefits from compact, spike-based communication between nodes instead of packets of data characteristic of traditional parallel algorithms wherein information is carried by the relative timing between *spikes* (timed all-or-nothing events) rather than numerical values passed between nodes. Memory bandwidth is the limiting computational resource in traditional large sparse linear solvers [8], and our approach colocalizes the computation and memory required to solve these problems to individual neurons and synapses.

**Spiking Neural Network for FEM problems**

Our approach begins by interpreting the solution process of the sparse FEM system ($\mathbf{A}x = b$) as a dynamical system which we embed into a spiking neural network (Figure 1B). We based this embedding on previous research in computational neuroscience that described a procedure for constructing spiking neural networks that implement a dynamical system [9]. This mapping of an FEM linear system to the spiking network is direct and therefore does not require any training during network construction. We focus initially on the steady-state Poisson equation on a disk with Dirichlet (i.e., fixed) boundary conditions (see Methods) as a prototypical example of a linear, elliptic PDE well suited to finite element methods, and often used as a benchmark problem. This problem also has an analytic solution allowing us to compare our spiking approximations to a ground truth. Our finite element formulation of the Poisson problem uses piecewise linear elements (Figure S1)

In our formulation (Figure 1), we associate a small population (typically between 8 and 16, but this is a free hyperparameter) of recurrently connected neurons with each mesh node. The sparse system matrix $\mathbf{A}$ determines the synaptic weights between neurons at different nodes (see Figure 1B and materials and methods). The right-hand side of the linear system manifests as biases applied to each neuron. Finally, each neuron receives an independent noise signal, which is important for achieving balanced, asynchronous spiking dynamics (see materials and methods).

Our neurons do not produce real-valued activations as output. Instead, the results of local computations in each neuron are communicated through *spikes*. In this sense, the network leverages spikes as *actions* that choreograph the activity of other neurons and the ultimate network output [10] as opposed to viewing spikes as single-bit numerical quantities or *symbols*. The spikes thus carry information through their relative timing relationships; thus, to solve numerical problems, the spike actions must be transduced to conventional binary representations of numerical quantities.

In our FEM network, the spikes from neurons representing a given mesh node are effectively low-pass filtered using a readout matrix $\mathbf{\Gamma}$ to construct an estimate of the solution of the FEM problem at that mesh node (Figure 1C). In other words, spikes act by perturbing a first-order linear dynamical system to drive it to the correct solution value (Figure 1C, bottom).

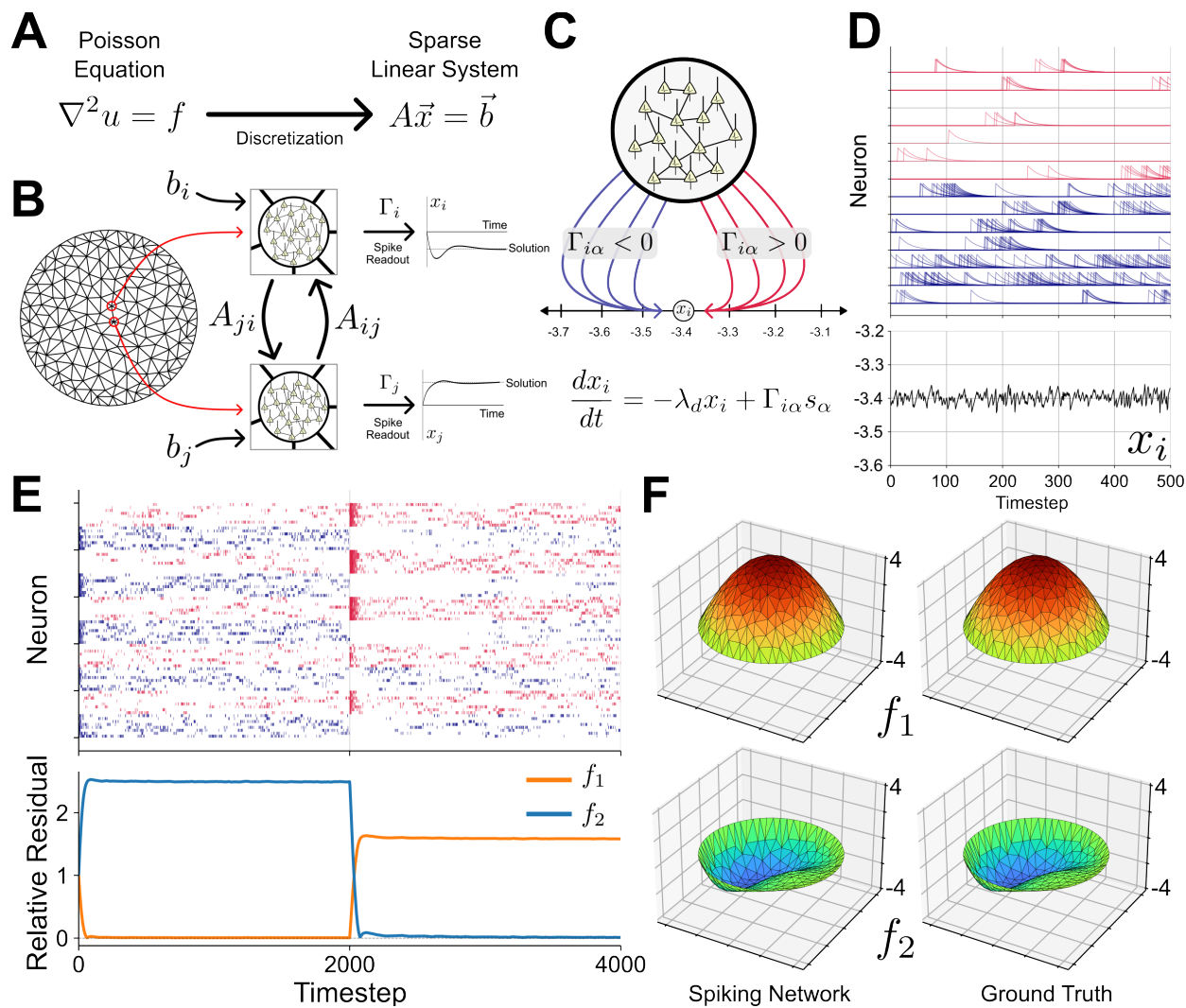

**Figure 1. Neuromorphic Finite Element Algorithm.** (**A**) Finite element methods convert partial differential equations (PDEs, such as the Poisson equation shown here) on a domain into a sparse linear system through a discretization process. The solution of the sparse linear system gives coefficients for approximating the solution of the original PDE. (**B**) The neuromorphic finite element algorithm (NeuroFEM) constructs a spiking neural network from the sparse linear system. Each mesh node in the discretization contains a population of spiking neurons. The weights between these neural populations are determined by the elements of the linear system matrix. Each population receives a bias determined by the right-hand side of the linear system. The network's construction ensures that the readout of the spiking activity flows to the solution of the linear system over time. (**C**) Spikes from neurons within a single mesh node are read out through a low-pass filtering process. Spikes project to output variables with weight Γ and are integrated through a first-order ordinary

differential equation (bottom). Half the neurons in a node project with a positive readout weight and half with a negative weight. This tug-of-war between positive and negative projecting neurons pulls the readout variable to the required value. (**D**) Spikes from individual neurons in a node contribute an exponentially decaying kernel to the readout variable. Red traces indicate kernels with positive readout weight and blue traces indicate negative readout weights. The sum of these kernels over all neurons in the node gives a fluctuating readout for the component of the linear system corresponding to this node ($x_i$). (**E**) NeuroFEM solves sparse linear systems. The activity of a NeuroFEM circuit constructed to solve the Poisson equation on a disk flows to the solutions with respect to two different right-hand sides $f_1$ and $f_2$ (top). Spiking activity from 5 nodes in the network is sparse and irregular. Red spikes are positive projecting neurons and blue spikes are negative projecting neurons. The network flows to a steady state corresponding to the solution of the linear system with respect to right-hand side $f_1$. At timestep 2000, the bias to each neuron was changed to reflect the new right-hand side $f_2$. The spiking activity immediately responds and settles into a new steady state that corresponds to the new solution (bottom). The relative residual with respect to $f_1$ of the instantaneous readout from the NeuroFEM circuit flows toward zero over time indicating that the neural readout solves the linear system. At timestep 2000, the network switches to solving the system with respect to $f_2$ and the relative residual with respect to $f_2$ again flows to zero. (**F**) Three-dimensional plots of the NeuroFEM-generated solutions (left) with respect to right-hand sides $f_1$ (top) and $f_2$ (bottom) show close agreement with the ground-truth solutions generated by conventional linear solvers (right).

The spiking dynamics of our spiking FEM network (referred to as NeuroFEM) are shown in Figure 1E, top. Over time, the network converges to a steady-state, asynchronous firing regime. Figure 1E, bottom shows the relative residual between the true solutions of the linear system (for two different forcing functions $f_1$ and $f_2$, see materials and methods) and the instantaneous readout of the spiking dynamics in the top panel, showing that the network's steady-state spiking well-approximates FEM solutions. Figure 1F shows the solution of the spiking network rendered onto the original finite element mesh, showing close agreement with classical solutions for the different forcing functions $f_1$ and $f_2$ (i.e., right-hand sides of the linear system).

We found that a direct translation of the networks prescribed by [9] does not yield a satisfactory linear solver, with such networks producing a steady-state bias in their solutions (Figure S2). We overcame this problem by realizing that [9] is mathematically equivalent to a proportional-only controller, which necessarily yields a steady-state bias. We fixed this by augmenting each neuron with an additional state variable that integrates the local residual error, turning the network into a system of distributed, spiking PI controllers (see materials and methods). This eliminates the steady-state error while still only requiring information local to each neuron.

Once a spiking network is constructed for a particular FEM problem (i.e., a choice of PDE, boundary conditions, mesh, and finite element space), the network may continuously solve new problem instances (right-hand sides; forcing functions) by varying the biases applied to each mesh node at run time. Figure 1E shows a switch to a new right-hand side representing a different forcing function at timestep 2000. The network immediately responds by adjusting its spiking dynamics to flow to a new steady state, representing the FEM solution of the PDE with respect to the new forcing function (Figure 1F, bottom). This reconfiguration is a consequence of the natural dynamics of the spiking network by construction and requires no training or adjustment beyond the initial construction of the FEM problem and the associated spiking network. This implies that a physical instantiation of this network could respond in real time to external data obtained from physical sensors, with a "neuromorphic twin" providing continuous estimates for the response of a real-world system—an important application domain for neuromorphic scientific computing [11].

Importantly, our NeuroFEM construction is scalable and adaptable to different mesh resolutions. Because the network architecture is determined by the sparsity structure of the system matrix **A**, the intrinsically geometric origin of this structure through the finite element mesh manifests as a geometrically structured spiking recurrent neural network with locally dense, globally sparse connectivity. Since neurons only synapse with their nearest neighbors in the mesh, the number of synapses per neuron remains essentially constant as the number of mesh nodes increases. This is critical because it allows our network to model complex geometries with large, unstructured meshes. Further, the accuracy of numerical solutions is tightly coupled to the resolution of mesh, and higher resolution meshes yield more accurate solutions (Figure 2B).

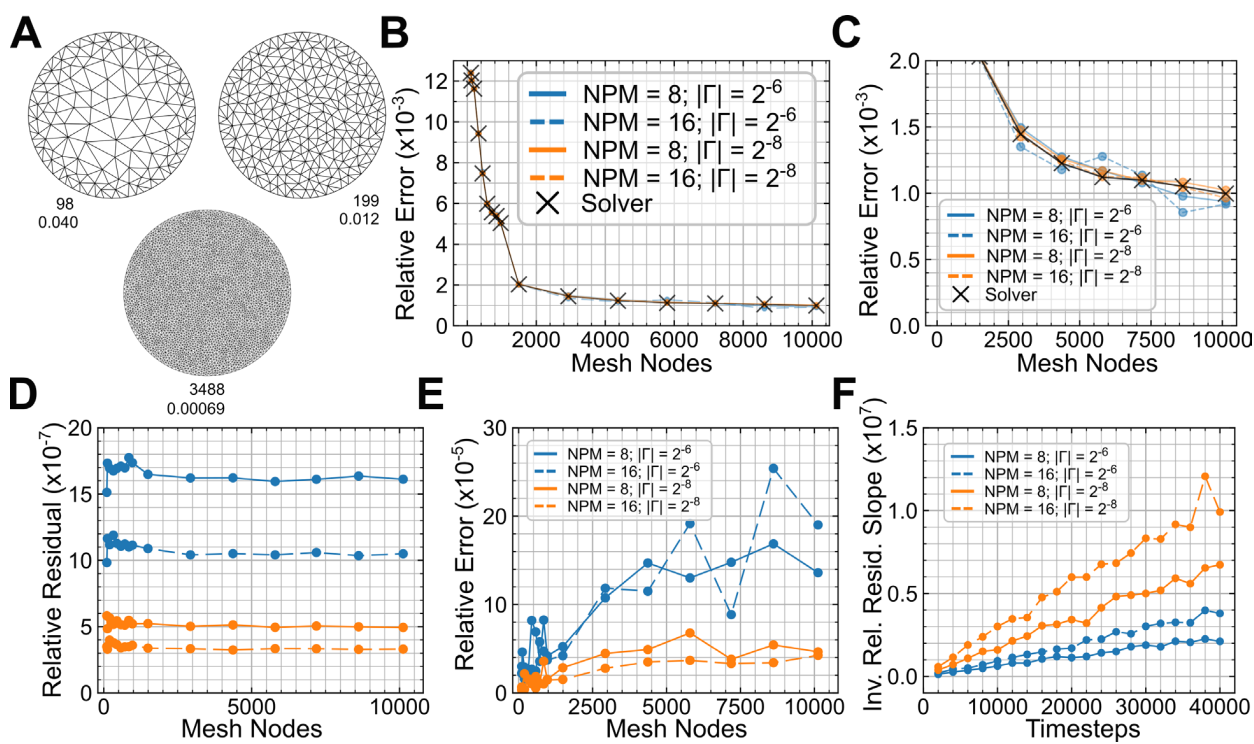

**Figure 2. Accuracy of NeuroFEM solutions**. **(A)** We evaluated NeuroFEM using the Poisson equation on a disk with Dirichlet boundary conditions as a benchmark problem. We increased the resolution of the mesh to assess how well NeuroFEM solutions converged to an analytic solution and compared to solutions generated by a conventional sparse linear solver (SciPy spsolve). Top numbers indicate the number of internal nodes and bottom numbers indicate the maximum triangle area in the mesh. **(B)** The relative error of different NeuroFEM solutions and a conventional solution with respect to the analytic solution as a function of the mesh resolution. As the number of mesh nodes increase, both NeuroFEM and conventional solutions approach the analytic solution. We tested NeuroFEM models with either 8 or 16 neurons per mesh node (NPM) and with a readout magnitude $|\Gamma|$ of either $2^{-6}$ or $2^{-8}$. All NeuroFEM models follow the trajectory of the conventional solver, and networks with smaller

|Γ| follow more closely. This is not a convergence study as the relevant quantity is the size of the circuit (number of nodes) rather than the element size. (**C**) Zoomed-in view of the curves in (B). (**D**) The relative residual of the linear system per mesh point for NeuroFEM is constant as a function of the number of mesh nodes. Colors and line styles are the same as (A) and (B). The relative residual is improved by increasing the number of neurons per mesh node and/or decreasing the readout magnitude |Γ|. (**E**) The relative error between NeuroFEM solutions and solutions generated by a conventional solver. As the mesh resolution increases, the error between NeuroFEM and the conventional solver asymptotes. (**F**) The relative residual improves linearly with the number of readout timesteps averaged together. Inverse relative residual slope is defined as the reciprocal of the constant relative residual per mesh node plotted in (D). Steeper curves indicate greater improvement as the number of averaging timesteps increases.

We examined the numerical accuracy of the NeuroFEM solutions for the steady-state Poisson equation on a disk, with Dirchlet boundary conditions and a constant forcing function across the domain, at several different mesh resolutions (approximately 100–10,000 nodes). Because the network produces a fluctuating estimate of the solution, these figures reflect the accuracy of the spiking readout averaged over 10,000 timesteps after the network reached steady state (see materials and methods). We observed that the spiking network provided comparably accurate solutions to those of a classic numerical solver (SciPy spsolve) as the number of mesh nodes increases (Figure 2, B and C). The relevant independent variable for Figure 2 is the size of the NeuroFEM circuit, implied by the number of mesh nodes. Convergence with respect to element size is a property of the specific finite element method, not the linear solver. For this example, NeuroFEM exhibits quadratic convergence (Figure S3) identical to the classical method, as expected for piecewise linear (order 1) elements (Figure S1).

We computed the relative residual of the linear system ($\frac{\|b - Ax\|}{\|b\|}$) and found that the relative residual per mesh node remained constant as the number of mesh nodes increased (Figure 2D), in agreement with conventional solvers. While the absolute value of this residual was worse for NeuroFEM than SciPy's spsolve we found that changing network parameters such as the number of neurons per mesh point (8 or 16) or the magnitude of the readout weights ($2^{-6}$ or $2^{-8}$) directly contributed to the residual magnitude, suggesting that our performance observations are due in part to particular parameter choices and not intrinsic limitations of the algorithm. This invites more detailed investigations of the numerical properties of the NeuroFEM solutions.

We directly compared the NeuroFEM solutions to the solutions provided by our classical solver (SciPy spsolve) by computing the relative error between the spsolve solution ($x$) and the NeuroFEM solution ($\hat{x}$) ($\frac{\|x - \hat{x}\|}{\|x\|}$) (Figure 2E). As the number of mesh nodes increased, this quantity appeared to reach an asymptotically constant value, indicating that the relative error between the spiking network solution and those from a conventional linear solver do not diverge as the size of the circuit increases.

Finally, we characterized the statistical properties of the fluctuating readout of the spiking network solution. Figure 2D shows that the slope of the relative residual as a function of the number of mesh nodes is constant for a particular network. We found that as we increased the number of samples we averaged to yield a solution, the inverse of this slope grew linearly (Figure 2F). In contrast, by adding noise to the classical solver solution, this quantity scaled as the square root of the number of samples (Figure S4). This is an important observation because it indicates that the spiking network output is *not* simply a noisy readout of an average solution (i.e., it is not a rate code). This has important implications for the ultimate efficiency of a neuromorphic solution because it implies that the number of spikes (and hence energy) of a neuromorphic circuit grows linearly, rather than quadratically, to reach a desired performance. [12]. The maximum standard deviation of the fluctuating readout across the mesh did not grow with network size (Figure S5)

**Implementation on Loihi 2**

The NeuroFEM circuit is broadly compatible with both analog and digital neuromorphic technologies. The Loihi 2 neuromorphic chip is a digital CMOS integrated circuit specifically designed for efficient evaluation of spiking neural networks with sparse connectivity and asynchronous spiking activity [13]. We examined the suitability of our spiking FEM algorithm on the Loihi 2 platform (Figure 3B), which enables up to 1 million neurons per chip and has been built into systems with over 1 billion neurons [14]. While our initial evaluations of the spiking network in Figs. 1 and 2 used floating-point CPU simulations, our specific Loihi 2 implementation must be tailored to account for the fixed-point precision of Loihi 2 hardware.

We implemented the NeuroFEM neurons (see materials and methods) using custom microcode written in Intel's proprietary Loihi 2 assembly language [15]. To convert our floating-point spiking networks to fixed-point, we rescaled our floating-point networks so that the parameters occupied fixed intervals bounded by powers of two, while preserving the overall dynamics of the networks (see materials and methods). Then, we rounded these parameters to fit the available bit widths on the Loihi 2 platform, such as the 8 bits of available synaptic weight precision and 24 bits of state variable precision. Our custom microcode neurons include instructions for bit shifting different parameters to common fixed-point representations, allowing us to choose appropriate scale factors for each parameter that maximize the available precision individually (i.e., different scale factors may be used for different parameters as needed). Nevertheless, optimizing the fixed-point conversion of our network is an important open question.

We evaluated FEM problems with different mesh resolutions (103–967 nodes) on a single Loihi 2 chip. We found that our spiking network on Loihi 2 was consistently able to approximate the solutions to the FEM linear system (Figure 3A, D and E). The relative residuals between the Loihi 2-generated solutions and the analytic solution were much larger than the floating-point or classical spsolve solutions but did not grow exceedingly with increasing mesh size (Figure 3, D and E). We found that networks with 16 neurons per mesh node and fewer than approximately 200 mesh nodes did not reliably converge, likely due to perturbations introduced by the fixed-point conversion and their outsized effect on such a small network. However, for larger meshes, all networks and runs converged to the same solutions. Thus, the fixed-point conversion necessarily introduces numerical errors in the spiking solution, but importantly, the fixed-point spiking network is stable on the hardware, at least for meshes with more than 200 nodes.

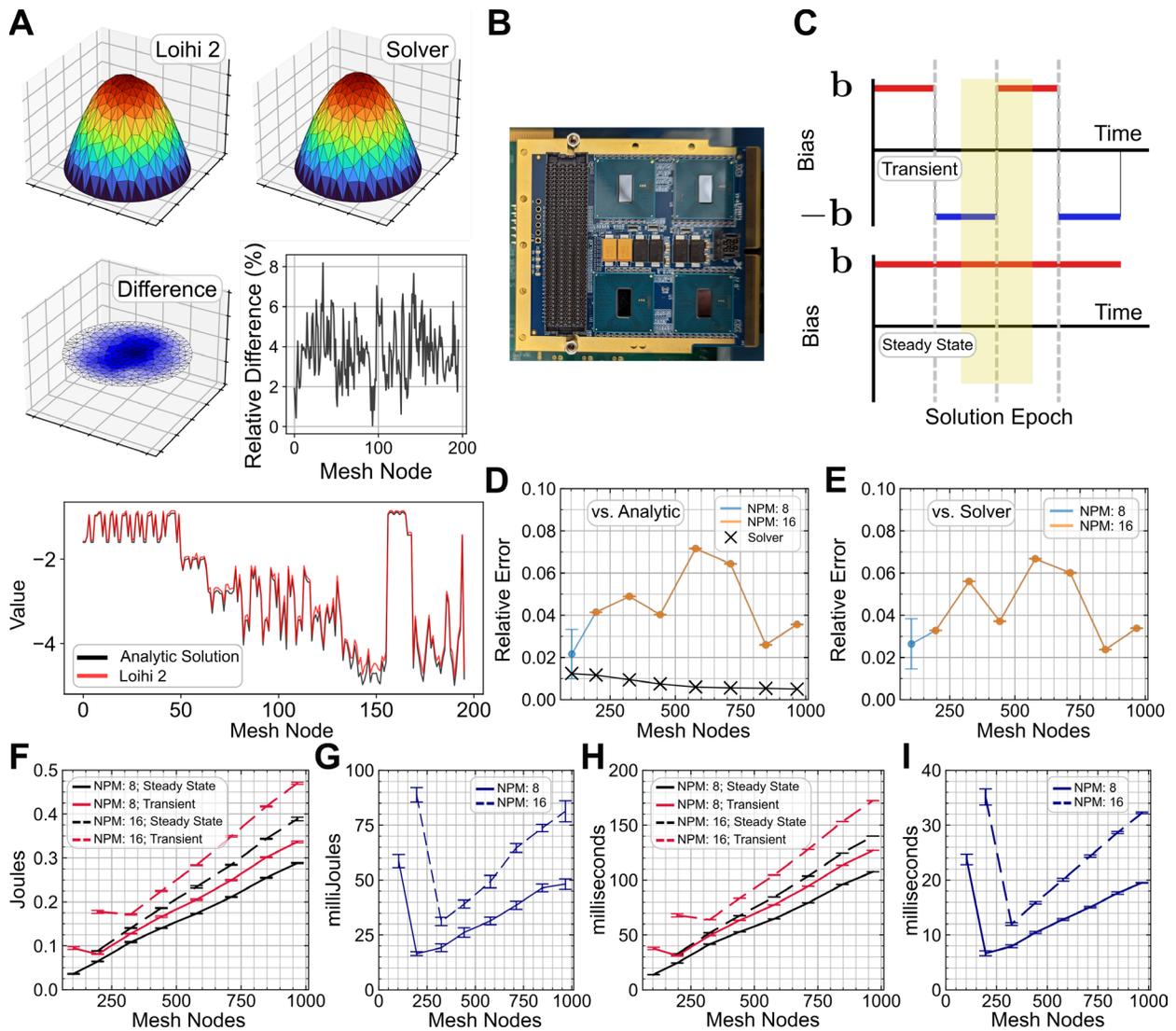

**Figure3. NeuroFEM on Loihi 2.** (**A**) (top) The NeuroFEM circuit on Loihi 2-generated solutions to the benchmark problem comparable to conventional solvers. Errors are likely due to the required fixed-point conversion (middle). The differences between the Loihi 2-generated solution and the conventional solution is localized to high-magnitude areas of the solution (left). The relative deviation between the Loihi 2 solution and the conventional solver was consistent across mesh nodes at around 4% on average and a maximum of approximately 8% (bottom). The Loihi 2-generated solution agrees in overall shape with the analytic solution, but large-magnitude values show greater deviation. (**B**) The Loihi 2 Oheo Gulch board used for this study. Each board has 8 chips (4 on the top and 4 on the bottom), and the full system has a stack of 4 boards. Our profiling studies were restricted to a single chip. (**C**) Schematic of the profiling strategy. NeuroFEM circuits on Loihi 2 were run in two modes: transient (top) and steady state (bottom). In the transient mode, the bias was sign-inverted every 4096 timesteps (a solution epoch) forcing the network to flow to a new solution. In the steady-state mode, the bias remained constant, keeping the network in its steady-state firing regime for the whole run. By comparing the energy and time difference between the two modes, we could estimate the energy and time required by each solution epoch. (**D**) The relative error with respect to the analytic solution for Loihi 2-generated solutions and conventional solutions. The fixed-point nature of Loihi 2 leads to worse performance compared to the CPU simulations of NeuroFEM, but the performance does not grow significantly in the range of mesh resolutions tested. The Loihi 2 implementation of the lowest-resolution mesh and 16 neurons per mesh node did not converge reliably, likely due to instabilities introduced by the fixed-point conversion and their outsized effect on such a small circuit. Error bars show the standard deviation over 10 separate runs. (**E**) The relative error between the Loihi 2-generated solutions and conventional solutions. The curves are largely similar to (D), because the conventional solutions are already close to the analytic solution. Error bars show the standard deviation over 10 separate runs. (**F**) Total energy required by each NeuroFEM model for 131,072 timesteps as a function of mesh resolution, neurons per mesh node, and run mode. Error bars indicate the standard deviation over 4 separate runs. (**G**) The energy difference per solution epoch for NeuroFEM models with either 8 or 16 neurons per mesh node. Error bars indicate the standard deviation. (**H**) The execution time per solution epoch as a function of mesh resolution, neurons per mesh node, and run mode. Error bars indicate the standard deviation. (**I**) The execution time difference per solution epoch for NeuroFEM models with either 8 or 16 neurons per mesh node. Error bars indicate the standard deviation.

**Energy and Time**

We next evaluated the energy and time requirements for solutions on Loihi 2. Because the real "work" of solving the FEM problem in our spiking network happens during the transient departures from steady-state firing, which we term "solution epochs" (Figure 1E, bottom and Figure 3C), we profiled the energy and time requirements by forcing the network to flip between sign-inverted right-hand sides every 4096 timesteps (see Figure 3C and materials and methods). We used Loihi 2's built-in power measurement systems to measure the energy and runtime difference between networks that flipped every 4096 timesteps to networks that remained in the steady-state firing regime for an equivalent total number of timesteps ($2^{17}$ = 131,072 timesteps; 32 solution epochs). This allowed us to estimate the additional energy and time required by each solution epoch.

We found that the required additional energy scaled approximately linearly as the number of mesh nodes increased (Figure 3, F and G). The required additional energy per solution epoch was at most approximately 80 milliJoules for meshes up to approximately 1000 nodes. Not surprisingly, doubling the number of neurons per mesh node increased the required energy, but interestingly, did not double it. Similarly, we found that the time difference per solution epoch scaled linearly with the number of mesh nodes (Figure 3, H and I), requiring a maximum of about 30 milliseconds longer per solution epoch for the transient case compared to the steady-state network for the meshes we considered. Iterative solvers on modern CPUs are highly optimized and their profiling is mesh-dependent and beyond the scope of this study, but we do note that while this same system can be solved faster on CPU with either GMRES (~1-2x faster) and conjugate gradient (~10x-20x faster), we estimate that the energy cost for a solution is significantly less on Loihi (Figure S6). Further, we expect that this energy advantage will grow with larger and more complex systems. Expanded details on this energy comparison are provided in the Supplementary Text.

We finally examined the scaling of the NeuroFEM algorithm on the full 32-chip Oheo Gulch Loihi 2 platform. In these experiments, we used all 32 chips and examined an increasing number of cores used per chip (Figure 4A). We first examined strong scaling, which describes how performance improves when adding more cores to a problem of fixed size (Figure 4B). We observed that up to a point, the NeuroFEM approach had nearly ideal strong scaling on Loihi 2 (i.e., doubling the number of cores halved the total time). Strong scaling always eventually saturates due to Amdahl's law, which dictates that there is a point at which costs that cannot be parallelized, such as communication and serial processing, dominate. When using an estimate of the serial cost (a single Loihi 2 core does not have sufficient memory to store the full model), we observe ideal strong scaling up to a factor of ~200 to ~600, depending on the model size. Per Amdahl's law, this confirms that the NeuroFEM algorithm is over 99% parallelizable.

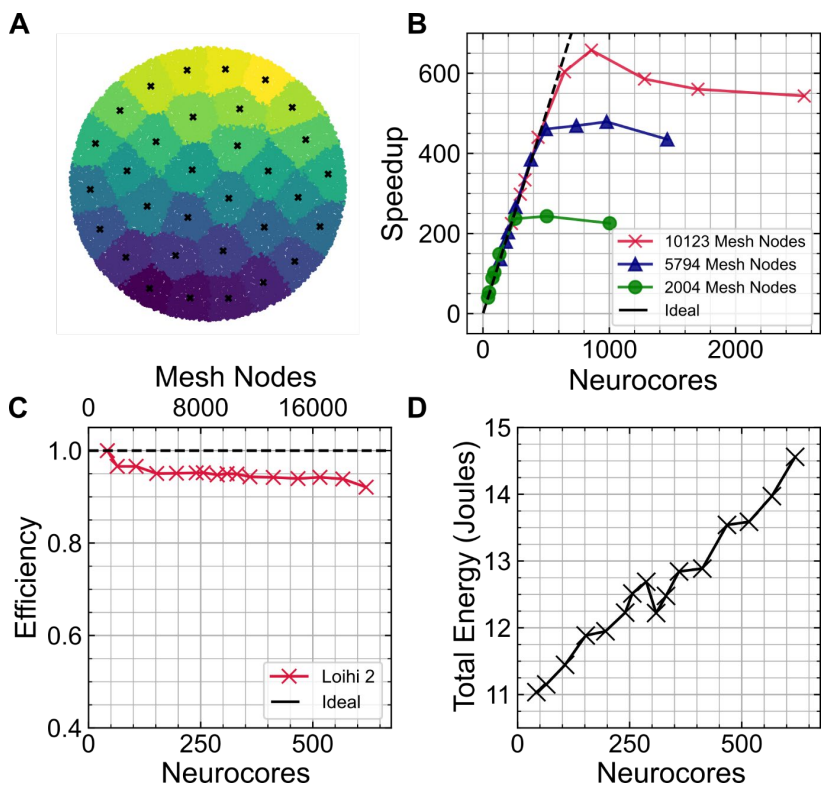

**Figure 4. NeuroFEM shows both effective strong- and weak-scaling when implemented in multi-chip Loihi 2 system.** (**A**) Illustration of multi-chip partitioning of 10,123 node mesh across 32 chips. Each color represents mesh points belonging to a given chip and each **x** is the centroid of mesh points on that chip. (**B**) Strong scaling of NeuroFEM shows ideal scaling performance up to several hundred cores for models of fixed size, after which scaling improvements cease due to Amdahl's law when model is too small to benefit from additional cores. (**C**) Weak scaling of NeuroFEM shows near ideal efficiency when model size increases with number of cores. (**D**) Weak scaling of multi-chip NeuroFEM results in a modest increase in total energy consumption at a rate lower than 1, suggesting increased energy efficiency at scale. All simulations were on all 32 chips for 131,072 timesteps.

We next examined weak scaling, which describes the case where the number of cores used grows with the problem size (Figure 4C). We observed that for a fixed number of mesh nodes per core (32), the NeuroFEM model shows near ideal weak scaling up to a point; highlighting that it is nearly as efficient to run a big model on many cores as a small model on fewer cores. However, as with strong scaling there is a limit to this efficiency. Notably, the increased number of cores used in the weak scaling experiment did modestly increase overall energy consumption linearly (Figure 4D), but at an overall slope lower than the slope of weak scaling on a single CPU (where energy and time should be proportional with constant power), suggesting that energy advantage for Loihi 2 may grow with increased model sizes (Figure S7) Together, these strong- and weak-scaling results suggests that this NeuroFEM formulation is effective up to the point of the specific neuromorphic architecture designs.

**Broader impact**

A significant benefit of this approach is that it enables the direct use of neuromorphic hardware on a broad class of numerical applications with almost no additional work for the user. If a problem can be represented as a sparse linear system, the neural circuit can directly implement the system. To illustrate this, we used NeuroFEM (on conventional CPUs) to solve more complex FEM problems than the 2D Poisson equation on a disk. First, we generated a more topologically complex 2D domain by putting holes in the disk (Figure 5A) and a more complex, unstructured mesh with spatially inhomogeneous resolution. To illustrate more complex boundary conditions, we formulated the problem with Dirichlet boundary conditions (i.e., fixed temperature) on the outer boundary and Neumann boundary conditions (i.e., fixed heat fluxes) on the inner hole boundaries. NeuroFEM successfully solved these systems (Figure 5B) and the generated solutions were close to conventionally generated solutions (Figure 5C).

Extending NeuroFEM even further, we solved a static linear elasticity problem in three dimensions (Figure 5D), namely, the deformation of a 3D shape under its own weight due to gravity with one face fixed. Unlike the 2D Poisson examples, this example features a more complex system of PDEs (see materials and methods) and a topologically nontrivial 3D tetrahedral mesh. Furthermore, the solution is a vector field representing the displacement of the shape (Figure 5E) rather than a scalar field. NeuroFEM again successfully solved this system with small deviation from a conventional solver (Figure 5F), a deviation which can be improved by further optimizations (Figure 2D). Importantly, this problem was generated by conventional FEM tools, Gmsh [12], and SfePy [16, 17] and translated directly to NeuroFEM. This means that neuromorphic algorithms can now interface with traditional scientific computing packages, dramatically reducing the barrier to entry for neuromorphic hardware.

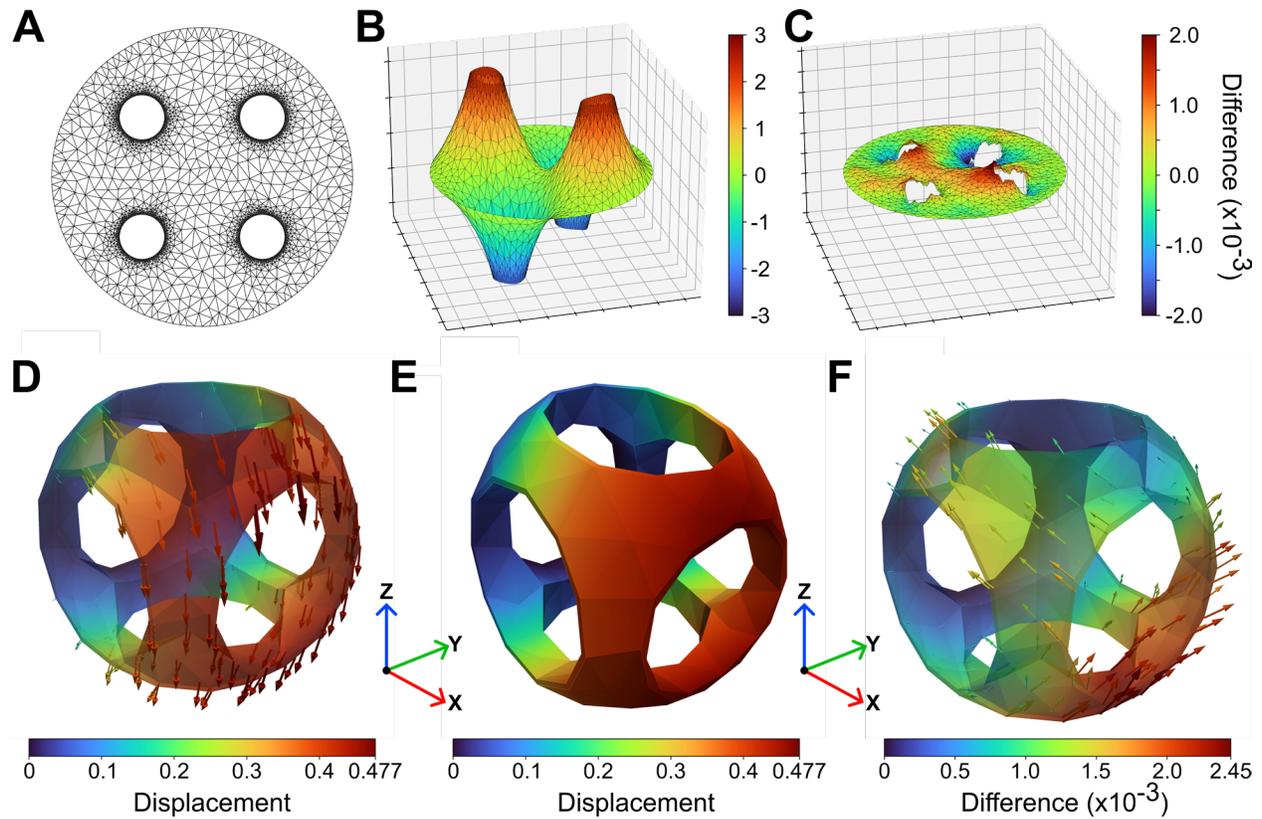

**Figure 5. NeuroFEM solves more complex FEM problems.** (**A**) Topologically nontrivial mesh with holes and spatially inhomogeneous mesh resolution. (**B**) NeuroFEM-generated solution for the Poisson equation on the topologically nontrivial mesh with Dirichlet boundary conditions on the outer boundary and Neumann boundary conditions on inner hole boundaries. (**C**) Difference between the NeuroFEM solution (B) and the conventional solution. Differences are bounded in magnitude by $2 \times 10^{-3}$. (**D**) NeuroFEM solution to the linear elasticity equations on a topologically nontrivial tetrahedral (3D) mesh subject to a constant volume force density (gravity) in the –Z direction. The –X face is fixed in position, and the rest of the faces are free. The solution is now a vector field instead of a scalar field. (**E**) Same as (D) but with the displacement vectors applied as a deformation of the original mesh. The solution agrees with the expectation of the hollow solid sagging under gravity. (**F**) The difference between the NeuroFEM solution and a conventional solution. The magnitudes of the difference vectors are bounded by $2.45 \times 10^{-3}$.

## Discussion

In this paper, we have demonstrated that a standard approach to numerically solving PDEs can be implemented directly on spiking neuromorphic architectures in a manner which effectively leverages the energy-efficient features of spiking hardware. Neuromorphic computing has been increasingly proposed as a solution to the growing energy crisis in computing. But, in large part because it is an entirely new paradigm for computing, there have been few specific examples in which neuromorphic computing has been shown to provide the requisite accuracy demanded by real-world applications. To date, most such successes have been shown in analog crossbars [4, 18], which can make dense linear algebra operations very efficient; however, most large-scale scientific computing tasks are sparse at scale and in this regime, data movement dominates energy costs [19]. The approach described here takes advantage of that sparsity through efficient spiking communication, reducing communication between mesh points to a minimal set of discrete spike events. While we have demonstrated this approach on fully digital CMOS spiking architectures, this approach should extend to future neuromorphic platforms that mix local analog computation with spiking communication over distance to maximize efficiency [20-22]. Since multi-chip implementations confer additional energy costs, the observed modest rise in energy costs under weak-scaling indicates that the energy benefits should be even more evident on higher density architectures, such as wafer-scale systems and hybrid analog / digital platforms.

This approach provides several notable benefits beyond energy efficiency. First, our approach is a direct mapping of an established numerical method to neuromorphic computing. As a result, neuromorphic computing will benefit from established numerical theory. Relatedly, unlike most neural algorithms, our approach is directly programmed and does not require training or learning. While the ability to learn from a naïve model underlies much of the value of machine learning methods, the ability to implement established models has value in terms of confidence and reliability.

Indeed, most neural approaches to scientific computing to date have focused on scientific machine learning (SciML) approaches, such as deep operator networks (DeepONets) and physics-informed neural networks (PINNs) [23-25]. While these techniques are promising as surrogate models, these methods often still require conventional simulations to provide training data, and these neural networks have largely been optimized for the use of GPUs and similar architectures. Although there are increased efforts to map these techniques onto neuromorphic systems [26], it remains an open question whether neuromorphic hardware can outperform GPUs on deep neural networks, which have largely evolved to benefit from GPU's single-instruction multiple data (SIMD) architecture [27]. The ability to take advantage of non-GPU co-processors is important for future computing systems as neuromorphic hardware in principle allows fully distributed algorithms and can tolerate heterogeneity naturally, as is evident with the complex geometries shown in Figure 5.

Finally, an important virtue of our approach is that it should place no additional work on the application developer to effectively utilize neuromorphic hardware. To date, while there have been several demonstrations of algorithms with neuromorphic advantages [6, 28-30], they largely require computational problems to be reformulated in a manner that is different from what is conventionally used in today's scientific computing (this is similarly the case for quantum computing and many other emerging computing paradigms [31-33]). For example, in our previous demonstration of simulating Monte Carlo random walks [6], the neuromorphic advantage arises when particles are virtually represented over a state space as opposed to a more conventional representation of allocating particles to processors. In such cases, it is necessary to also account for the user cost in considering problems from a new perspective, which ultimately increases the cost of adoption. In the approach described here, without modification we can import existing finite element mathematics wholesale to the neuromorphic domain. Thus, neuromorphic hardware can now directly benefit from the decades of CAD/FEM assembly tools developed to enable numerical simulation; likewise, this allows the established numerical computing community to readily examine neuromorphic hardware. The user-unfriendliness of spiking neuromorphic hardware has long been recognized as a significant limitation to broader adoption [1] and our results directly mitigate this problem.

There is a final advantage of this approach that merits further exploration. The original neurobiological motivation for this class of spiking neural networks was to model cortical motor control circuits [9]. While our implementation is not a direct mapping of this original approach, the success of numerically solving PDEs in a related network suggests that cortical-like neural circuits are indeed capable of similar computations. While it is unlikely that the cortex solves PDEs directly in this way, it is not unreasonable that the underlying neural dynamics and their ability to provide online estimations of the physics of the world may be relevant for understanding higher level cortical computations.

**Acknowledgments:** We thank Craig M. Vineyard and the DOE NNSA Advanced Simulation and Computing program for access to the Loihi 2 platforms used in this study and helpful discussions. We acknowledge J. Darby Smith, Michael Krygier, Felix Wang, William Severa, John Shadid, Fred Rothganger, Rasmus Tamstorf, Erik Boman, and John Wagner for helpful discussions and we thank Lindsey Aimone for scientific copyediting. We would like to thank the Intel Neuromorphic Research Laboratory for providing documentation and advice for programming and profiling Loihi 2.

This paper describes objective technical results and analysis. Any subjective views or opinions that might be expressed in the paper do not necessarily represent the views of the U.S. Department of Energy or the United States Government. Sandia National Laboratories is a multimission laboratory managed and operated by National Technology & Engineering Solutions of Sandia, LLC, a wholly owned subsidiary of Honeywell International Inc., for the U.S. Department of Energy's National Nuclear Security Administration under contract DE-NA0003525.

**Funding:** We acknowledge support from the DOE Office of Science (ASCR / BES) Co-Design in Microelectronics project COINFLIPS.

## Methods

Finite Element Problem

We evaluated our FEM network on a specific PDE problem, the steady-state Poisson equation on a disk, with Dirchlet boundary conditions:

$$\nabla^2 u = f \text{ on } \Omega$$
$$u = 0 \text{ on } \partial\Omega.$$

Unstructured finite element meshes were generated with the MeshPy library, which provides a Python interface to the Triangle package (*34*). We generated meshes with different resolutions by setting the maximum triangle area parameter in the mesh generation process.

We assembled sparse linear systems (the stiffness matrix and mass matrix) using piecewise linear elements directly from their definitions using the NumPy and SciPy libraries.

For our accuracy and convergence evaluations (Figs. 2 and 3), we used a constant forcing function

$$f(x,y) = -20.$$

For this specific $f$, the problem has an analytic solution:

$$u(x,y) = 5(1 - x^2 - y^2).$$

For Figure 1E, we evaluated the solution with a different forcing function defined by

$$f(x,y) = 12 - 60(x - 0.25)^2 - 60(y + 0.13)^2.$$

Discretization of these problems on our meshes with piecewise linear elements yields a sparse linear system $Ax = b$ where $A$ is an ($n_{mesh}$, $n_{mesh}$) sparse matrix.

Spiking Neural Network

Our spiking neural network consists of neurons implementing generalized leaky integrate-and-fire dynamics. Readout neurons implement leaky integration of spikes $s_\alpha$ into real-valued output variables $x_i$ through a readout kernel $\Gamma$ through the differential equation:

$$\frac{dx_i}{dt} = -\lambda_d x_i + \sum_\alpha \Gamma_{i\alpha} s_\alpha(t).$$

$\boldsymbol{\Gamma}$ is our readout matrix which maps individual neurons to the variables of the linear system. For each mesh point, we associate a fixed quantity of neurons (*npm*; neurons per mesh point); thus, $\boldsymbol{\Gamma}$ has shape $n_{mesh} \times (n_{mesh} * npm)$. Nonzero elements in each row *i* of $\boldsymbol{\Gamma}$ correspond to neurons that project to the associated output variable $x_i$. Thus, there are *npm* nonzero elements per row. Half of these nonzero elements have magnitude $+|\Gamma|$, corresponding to neurons projecting with positive weight (Figure 1C) and half have magnitude $-|\Gamma|$. Thus, $\boldsymbol{\Gamma}$ has the form

$$\begin{bmatrix} +|\Gamma| & -|\Gamma| & 0 & 0 & 0 & 0 \\ 0 & 0 & +|\Gamma| & -|\Gamma| & 0 & 0 \\ 0 & 0 & 0 & 0 & +|\Gamma| & -|\Gamma| \end{bmatrix},$$

where each nonzero block is a $(1 \times npm/2)$ constant block with the indicated values.

There are two classes of synapses between neurons. 'Slow' synapses communicate information between mesh nodes and are called slow because they are twice integrated into the neuron's membrane potential. The slow synaptic weights are defined by

$$\boldsymbol{\Omega}_{slow} = \boldsymbol{\Gamma}^T \boldsymbol{A} \boldsymbol{\Gamma},$$

where $\boldsymbol{\Gamma}$ is the readout matrix, and $\boldsymbol{A}$ is the system matrix. $\boldsymbol{\Omega}_{slow}$ is block structured and sparse. Each nonzero block of $\boldsymbol{\Omega}_{slow}$ corresponds to a nonzero element of $\boldsymbol{A}$. Each nonzero block of $\boldsymbol{\Omega}_{slow}$ has the structure:

$$A_{ij} \begin{bmatrix} |\Gamma|^2 & -|\Gamma|^2 \\ -|\Gamma|^2 & |\Gamma|^2 \end{bmatrix}.$$

The entries in the 2 × 2 matrix above represent constant submatrices with shape $(npm/2) \times (npm/2)$. Even though $\boldsymbol{\Omega}_{slow}$ is already sparse, there is still significant redundancy in its definition and future implementations of the network may benefit from reducing this redundancy using shared synapses.

The second class of synapses are called 'fast' because they are singly integrated. They communicate entirely within mesh nodes and accomplish fast coordination between neurons within a node. The fast synaptic weight matrix is defined as

$$\boldsymbol{\Omega}_{fast} = \boldsymbol{\Gamma}^T \boldsymbol{\Gamma}.$$

Because of $\boldsymbol{\Gamma}$'s structure, $\boldsymbol{\Omega}_{fast}$ is block diagonal and each block has the form:

$$\begin{bmatrix} |\Gamma|^2 & -|\Gamma|^2 \\ -|\Gamma|^2 & |\Gamma|^2 \end{bmatrix}.$$

Slow synaptic input is first integrated into a latent state variable within each neuron ($u_1$) with time constant $\lambda_d$, according to the equation:

$$\frac{du_{1,\alpha}}{dt} = -\lambda_d u_{1,\alpha} + \sum_\beta \mathbf{\Omega}_{slow,\alpha\beta} s_\beta(t).$$

Here, $s(t)$ represents the vector of Dirac delta spike trains from other neurons in the network. Fast synapses are similarly singly integrated into a separate latent variable $u_2$ with the equation

$$\frac{du_{2,\alpha}}{dt} = -\lambda_d u_{2,\alpha} + \lambda_d \sum_\beta \mathbf{\Omega}_{fast,\alpha\beta} s_\beta(t),$$

where $u_1$ represents the neuron's local evaluation of the left-hand side of the linear system ($\mathbf{A}\mathbf{x}$). A neuron's component of the difference between the left-hand side and right-hand side of the linear system (the residual) is computed locally and carried by the latent variable $u_{err}$. Here, $\mathbf{b}$ is the right-hand side of the linear system and acts as a bias current for each neuron:

$$u_{err,\alpha} = -u_{1,\alpha} + \mathbf{\Gamma}^T b.$$

Each neuron acts as a PI controller (and integral dynamics are necessary for accurate spiking solutions), so each neuron has a state variable $u_{int}$ that tracks the integral of the error:

$$\frac{du_{int,\alpha}}{dt} = u_{err,\alpha}.$$

Finally, all these quantities are combined into a differential equation for the membrane potential $v$:

$$\frac{dv_\alpha}{dt} = -\lambda_v v_\alpha + k_p u_{err,\alpha} + k_i u_{int,\alpha} + u_{2,\alpha}$$
$$- \sum_\beta \Omega_{fast,\alpha\beta} s_\beta(t) + \sigma_v \eta.$$

Here, $\eta$ is a vector of independent samples from a Gaussian distribution. The parameter $\sigma_v$ is the standard deviation of the noise, which we set to 0.00225 for all networks.

Neurons emit a spike when their membrane potential $v$ reaches a threshold value $\theta$. The threshold is defined by

$$\theta = \frac{1}{2}|\Gamma|^2.$$

Upon spiking, neuron membrane potentials are reset by subtracting the threshold value.

## CPU Simulations

We simulated the above equations using Python. The above differential equations were integrated using the forward Euler method with a timestep $dt = 2^{-12}$.

## Spiking Network Evaluation

We evaluated our spiking network by comparing spiking solutions to both the analytic solution for our test problem, and to solutions produced by a conventional linear solver (SciPy spsolve function). We ran each network for 50,000 timesteps to ensure the network converged to steady state. We then extracted the spiking network's estimate of the solution by averaging the readout variables over the last 10,000 timesteps.

For all error estimates, we use the L2 norm. We computed the relative error between the analytic solution on the mesh points and either the spiking or conventional solver using the equation:

$$RelErr = \frac{\|\hat{x} - x\|}{\|x\|}.$$

Here, $\hat{x}$ represents either the spiking or conventional solution and $x$ represents the analytic solution.

We evaluated the relative residual of the linear system per mesh node (Figure 2D) for the spiking network by computing the quantity:

$$r = \frac{1}{N_{mesh}} \frac{\|b - A\hat{x}\|}{\|b\|}.$$

## Loihi 2 Implementation

We implemented the dynamical systems defining each neuron described by the equations above using custom microcode neurons on Loihi 2.

To map the parameters of our spiking network onto Loihi 2, we first converted the above equations from floating point to fixed point. To do this we rescaled the above equations [as described in (9)] so that the numerical values of the fast and slow weight matrices were contained in intervals bounded by a power of two (arbitrarily chosen).

Next, we introduced power-of-two scale factors for each parameter and state variable. Generically, if $x$ refers to a parameter or state variable in the floating-point model, we relate $x$ to its fixed-point counterpart (denoted $\bar{x}$) using the relation:

$$\bar{x} = x * 2^{S_x}.$$

Fixed-point quantities denoted with overbars are rounded to the nearest integer. After defining the associated fixed-point variables for each floating-point variable, we substituted the corresponding fixed-point quantity into the above equations. This substitution brought the associated scale factors into the above differential equations. We algebraically rearranged these scale factors so that the left-hand sides of each equation were written entirely in terms of fixed-point quantities.

This algebraic rearrangement leads to power-of-two scale factors multiplying each quantity on the right-hand sides of the equations. These factors are the conversion factors between the scales of the different fixed-point quantities in the equations. Because these conversion factors are all powers of two by definition, we realized these conversions on Loihi 2 by introducing specific bit shift instructions into our microcode neurons.

We chose the exponents in the scale factors by considering the allowed precision of different variables on Loihi 2. For example, synaptic weights on Loihi 2 have 8-bit precision. Thus, we chose the scale factor so that our final weight matrix would take the values in the range [–128, 127]. Similarly, state variables have 24 bits of precision, so we chose scale factors for each dynamical variable such that the dynamic range of that variable used as much of the available dynamic range of the corresponding Loihi 2 fixed-point variable. Once scale factors were chosen, the fixed-point equations determined the precise form of the microcode with all the required bit shifts to rescale variables to compatible fixed-point representations.

Loihi 2's pseudorandom number generators produce uniformly distributed 24-bit integers. Because the floating-point model assumed a Gaussian noise source with a specified standard deviation, we scaled the Loihi 2-generated random number so that the standard deviation of the uniform random numbers approximated that of the Gaussian noise source in the floating-point model, written in the scale of the neuron membrane potential.

| Parameter | Scale Factor |
|---|---|
| $\Omega_s$ | $2^6$ |
| $\Omega_f$ | $2^{19}$ |
| $\Gamma$ | $2^{13}$ | 
| v | $2^{28}$ |
| bias ($\Gamma^T b$) | $2^{17}$ |
| $u_1$ | $2^{16}$ |
| $u_2$ | $2^{20}$ |
| $u_{err}$ | $2^{16}$ |
| $u_{int}$ | $2^{20}$ |
| x (readout) | $2^{16}$ |

We chose the parameters $K_p$, $K_i$, $\lambda_d$, $\lambda_v$, and $dt$ to be powers of two, so that multiplication by these parameters becomes simple bit shifts in the fixed-point implementations. These bit shifts were folded into the shifts resulting from differences in fixed-point scale factors.

| Parameter | Value |
|---|---|
| $K_p$ | $2^2$ |
| $K_i$ | $2^4$ |
| $\lambda_d$ | $2^3$ |
| $\lambda_v$ | $2^4$ |
| $dt$ | $2^{-12}$ |

In Loihi 2, spikes from adjacent neurons are accumulated after weight multiplication into registers called dendritic accumulators. In our network, each neuron uses two dendritic accumulators to accumulate the slow and fast synaptic inputs separately. We refer to these accumulators as $DA_{slow}$ and $DA_{fast}$. The final fixed-point equations defining Loihi 2 neuron updates on each timestep ($n$) are as follows (here, >> indicates arithmetic right shift and << indicates arithmetic left shift):

$$\overline{u_1}[n+1] = (511 * \overline{u_1}[n]) \gg 9 + DA_{slow} \ll 10;$$

$$\overline{u_2}[n+1] = (511 * \overline{u_2}[n]) \gg 9 + DA_{fast} \gg 5;$$

$$\overline{u_{err}}[n+1] = \overline{u_1}[n] + \left(\overline{\Gamma^T b}\right) \ll 7;$$

$$\overline{u_{int}}[n+1] = \overline{u_{int}}[n] + (\overline{u_{err}}[n] \gg 8);$$

$$\overline{v}[n+1] = (255 * \overline{v}[n]) \gg 8 + (\overline{u_{err}} \ll 2) + \overline{u_{int}} + (\overline{u_2} \gg 4) - (DA_{fast} \ll 9) + (\overline{\eta} \gg 3).$$

Loihi 2 chips contain 128 neural cores. We mapped our network onto these cores by placing the neurons belonging to a single mesh node onto the same core, proceeding core-by-core and node-by-node (round robin). We wrapped the node to core mapping once we hit the maximum number of cores per chip, so the core index $c$ containing neurons for mesh node index $m$ is given by $m = c$ mod 128. This is a naïve layout of the network on the chip, and the network could benefit from a layout that respected the geometric adjacency relations between mesh nodes.

Loihi 2 Evaluation

We ran NeuroFEM circuits for mesh sizes in the range (100–1000) to ensure the circuits fit onto a single Loihi 2 chip. For each mesh resolution, we ran the circuits for 10,000 timesteps and computed the average value of the readout over the last 1000 timesteps. To quantify the accuracy of the Loihi 2 generated solutions, we computed the relative errors and residuals using the same formulae described above for the floating-point CPU simulations.

Loihi 2 Profiling

To profile the networks on Loihi 2, we used the same meshes as in our evaluations, but we turned off all I/O so that I/O was excluded from power and time measurements. Because the dynamics of the circuit relax to a steady-state solution, the real "work" in solving the linear system happens during transients in the execution upon a change in the bias (right-hand side). To profile these transients, we set up our profiling runs to alternate between two different right-hand sides by flipping the sign of the bias every 4096 timesteps. Because the system is linear, the resulting solutions are sign-inverted mirror images of each other. This means that during the solution transient, the activity in the network shifts from one set of neurons to the complementary set of neurons.

We performed two sets of profiling runs, which we term "steady state" and "transient." During steady-state runs, the biases were multiplied by +1 every 4096 timesteps, so the biases did not change, and the network remained in its steady-state firing regime. During transient runs, we multiplied the bias by –1 every 4096 timesteps. Thus, the network was forced to flow to a new steady state. We implemented these multiplications directly in the microcode, allowing the network to update its own bias without any I/O requirements, and ensuring that both steady-state and transient runs executed the same instructions. We call the 4096-timestep windows around each sign inversion a "solution epoch".

Each run, either transient or steady-state, lasted for 32 solution epochs or 131,072 timesteps. We used Loihi 2's built-in power measurement interface to record the power/energy used by the entire run. Then, we divided this total energy/wall time consumed by the number of solution epochs to get an estimate for the amount of energy/wall time required by each solution epoch on average.

By comparing the energy per epoch from both transient and steady-state runs, we were able to estimate the average excess energy and time required by the solution transients as the network computed a new solution (Figure 3). Comparing against networks that remained in steady state allowed us to subtract the steady-state power.

Multi-chip Scaling on Loihi 2

For multi-chip studies, we used all 32 chips on the Intel Loihi 2 platform while varying the number of cores used per chip and the number of meshpoints $n_{mesh}$.

To partition our mesh across chips, we used a greedy heuristic scheme that, while not guaranteed to provide an optimal partitioning, was effective at reducing inter-chip communication. Using our 2-d mesh as an example, this approach worked as follows:
- Initially, each chip is assigned an initial position $(x_0, y_0)$
- On each iteration, $k$,
    - Each mesh point is assigned to the nearest chip according to a weighted distance measure that accounts for load balancing,
    - Each chip's position $(x_k, y_k)$, is updated to be the centroid of its mesh points,
    - If a chip has no mesh points allocated to it, it is relocated to the position of the mesh point closest to it (and that mesh point is reassigned accordingly).

This partitioning is run for 16 iterations. While there is no guarantee of stability or speed of convergence, we observed that for 32 chips the partitioning was stable for our purposes (Figure 4A). Within each chip, mesh points were assigned to available cores in a round robin fashion.

For strong-scaling experiments, we examined the impact of using more cores per chip for several fixed model sizes. Strong-scaling studies typically seek to identify a measure of speedup,

$$speedup = \frac{T(1)}{T(N)} = \frac{1}{s + p/N},$$

where $N$ is the number of processors, $T(1)$ is the time for running on a single processor, $T(N)$ is the time for running on $N$ processors, and $s$ and $p$ are the serial and parallel proportions of the algorithm, respectively. In ideal strong scaling, $p>>s$, and there is a near-linear speedup for increasing $N$ up to the point that serial computation dominates.

In our experiments, because a single Loihi 2 core is limited to ~512 neurons, there was a minimum number of cores required for every experiment. This makes $T(1)$ impossible to measure empirically, so in our experiments we assumed that if core memory did not exist, we would see ideal scaling down to serial levels. This linear assumption is justified by the linearity of the first few points in our results and the agreement between scaling curves for meshes of different sizes.

For weak-scaling experiments, we examined the impact of increasing the model size with an increased number of model cores. As with the strong-scaling studies, we measured efficiency by the ratio of a single-core sized model against increasing sized models.

$$efficiency = \frac{T(1)}{T(N)},$$

In weak-scaling experiments, the ideal behavior is for time to be constant with increased model size since both size and resources are scaling together.

CPU Profiling

CPU profiling of GMRES (*35*) and CG (*36*) conventional linear solvers was performed using standard SciPy implementations on Python. These analyses were performed on an Intel Xeon E5-2665 server class processor. Full details on this processor are available here: (https://www.intel.com/content/www/us/en/products/sku/64597/intel-xeon-processor-e52665-20m-cache-2-40-ghz-8-00-gts-intel-qpi/specifications.html?wapkw=e5-2665)

Linear Elasticity Example

In Figure 5, D to F, we evaluated a linear elasticity problem on a 3D tetrahedral mesh. The mesh was generated using the Gmsh package (*12*) and consisted of 583 tetrahedra. The sparse linear system was assembled using SfePy (*16, 17*) and yielded a (3627, 3627) sparse matrix with 222,003 nonzero entries. We solved the resulting linear system using both SciPy spsolve and NeuroFEM using 8 neurons per mesh node (network size: 29,016 neurons) and $|\Gamma| = 2^{-6}$. Specifically, we solved the following PDE system (in Cartesian coordinates):

Cauchy momentum equation
$$\nabla \cdot \boldsymbol{\sigma} + \boldsymbol{F} = \boldsymbol{0};$$

Tensorial Hooke's law
$$\boldsymbol{\sigma}_{ij} = \sum_{kl} C_{ijkl} \boldsymbol{\varepsilon}_{kl};$$

Definition of infinitesimal strain
$$\boldsymbol{\varepsilon} = \frac{1}{2}[\nabla \boldsymbol{u} + (\nabla \boldsymbol{u})^T];$$

Definition of the stiffness tensor for a homogeneous, isotropic material
$$C_{ijkl} = \lambda \delta_{ij} \delta_{kl} + \mu (\delta_{ik} \delta_{jl} + \delta_{il} \delta_{kj});$$

where $\boldsymbol{\sigma}$ is the Cauchy stress tensor, $\boldsymbol{\varepsilon}$ is the strain tensor, $\boldsymbol{u}$ is the displacement vector, and $C$ is the fourth-order stiffness tensor. $\boldsymbol{F}$ is the force density (force per unit volume), which was constant in the $-\boldsymbol{Z}$ direction representing the force of gravity. The parameters $\lambda$ and $\mu$ are the Lamé parameters defining the material mechanical properties (both set to 1), and $\delta$ is the Kronecker tensor.

**Supplementary Text**

Connection between PDEs and sparse linear systems
We quickly review the weak formulation of PDEs and how this leads to sparse linear systems.
Consider the PDE
$$\nabla^2 u = f.$$

Multiply both sides by test functions $\phi_i$ and integrate over the domain $\Omega$ as follows:

$$\int_\Omega \phi_i \nabla^2 u \, d\Omega = \int_\Omega \phi_i \, f d\Omega.$$

In our case, the test functions $\phi$ are piecewise linear (Figure S4). Integrate the left-hand side by parts to yield

$$\int_{\partial\Omega} \phi_i \nabla u \cdot \hat{n} dS - \int_\Omega \nabla\phi_i \cdot \nabla u d\Omega = \int_\Omega \phi_i f \, d\Omega.$$

The first term on the left is a boundary term. If we impose Dirichlet boundary conditions, the $\phi_i$ are identically zero on $\partial\Omega$, so this term disappears.
We are left with

$$-\int_\Omega \nabla\phi_i \cdot \nabla u d\Omega = \int_\Omega \phi_i f \, d\Omega.$$

Now, we assume that $u$ is a linear combination of test functions:

$$u = \sum_j u_j \phi_j.$$

Substitution into the above equation yields

$$-\sum_j u_j \int_\Omega \nabla\phi_i \cdot \nabla\phi_j d\Omega = \int_\Omega \phi_i f \, d\Omega.$$

The integral on the left-hand side is nonzero only for elements that overlap. It depends on two indices, $i$ and $j$, so we can interpret this as a matrix. The right-hand side depends on index $i$, so we can interpret it as a vector. The result is a linear system of equations for the coefficients $u_j$:

$$A_{ij} = -\int_\Omega \nabla\phi_i \cdot \nabla\phi_j d\Omega;$$

$$b_i = \int_\Omega \phi_i f \, d\Omega;$$

$$\sum_j A_{ij} u_j = b_i.$$

Thus, we obtain a linear system for the coefficients $u_j$. Because the test functions are compactly supported, the only nonzero elements of $A_{ij}$ correspond to overlapping test functions. Thus, $A_{ij}$ inherits sparsity from the structure of the mesh.

Estimating the energy advantage of neuromorphic hardware

To estimate the potential energy advantage, it is necessary to compare the NeuroFEM algorithm on Loihi 2 to a conventional linear solver on a conventional hardware platform. There has been considerable research over past decades in optimizing linear solvers for FEM systems. For this reason, we focused on two standard linear solvers: GMRES (*35*) and conjugate gradient (*36*) in SciPy. These methods have different benefits and are suitable for different classes of FEM problems, but they provide a useful baseline to estimate time scaling and energy scaling of conventional linear solvers.

Additionally, while neuromorphic chips are widely recognized as being several of orders of magnitude lower power (~1 W for Loihi 2) than conventional processors (>100 W for a CPU, more for GPUs), the compute cores on neuromorphic chips are typically slower and parallel in operation. For this reason, a direct power comparison is not appropriate, rather an energy comparison is required. We evaluated conventional linear solvers on a server class Intel Xeon E5-2665 CPU, which has a 2.40GHz clock speed and a total draw power of 115 Watts (https://www.intel.com/content/www/us/en/products/sku/64597/intel-xeon-processor-e52665-20m-cache-2-40-ghz-8-00-gts-intel-qpi/specifications.html?wapkw=e5-2665).

Figure S6 shows scaling time scaling results of the conventional solvers on CPU that correspond to the same models shown in Figure 3. Like NeuroFEM, GMRES and CG scale linearly (as expected) with size. GMRES is ~10x slower, requiring ~50 ms to converge, while CG converges in ~5ms. Assuming ~100 W for the CPU power draw, we can estimate about 5 J for GMRES and 0.5 J for CG for the 1000 mesh node solution.

Obtaining a direct measurement for the NeuroFEM solution is not straightforward either, though we do use Intel Loihi 2's energy probes to begin our estimate. Because the NeuroFEM implement runs for a fixed time, rather than halting at a desired convergence, our experiments overestimate the overall time and energy required to solve the problem. Nevertheless, when the model is run for this overestimated number of timesteps, we still observe an *upper-bound* of energy consumption that is still less than CG (~0.3 J for 8 neurons per mesh node, 1000 mesh nodes).

To account for this overestimation of simulation time (and thus energy costs), we incorporated a transient in the simulation of a steady-state model (Figure 3C), observing that there is an overall increase in power and time (Figure 3F), we can attribute to the model 'solving' the problem. This solution is a *lower-bound* of overall energy consumption (~50 mJ) that is a factor of roughly 10x lower than CG.

Finally, we investigated the scaling of larger models on CPU that we examined in our weak scaling experiment (Figure 4C & D). Notably, this CPU experiment is on a single chip, which should maximize energy-efficiency, whereas the Loihi 2 weak scaling study was on 32 chips. Nevertheless, we observe that the overall time to simulate the largest 20,000 mesh node model on a single GPU 200 ms with CG, suggesting that ~20 J would be required, and at this size CG on the CPU is beginning to experience unfavorable scaling. This is still greater than the upper bound of the multi-chip Loihi 2 cost of ~15 J.

There are several caveats to this analysis, making it arguably preferable to rely more on the scaling results rather than direct comparisons. When possible, we make assumptions that favor the baseline (CPU) implementation. Important caveats are as follows:

1) It is difficult to directly measure the energy of modern CPUs chips. While we would expect that most of the draw power goes to the computationally heavy workload such as these, it is difficult to profile exactly.
2) Modern linear solvers and CPU compilation tools are highly optimized, whereas the neuromorphic hardware is still a research platform. For this reason, we focused on Python SciPy instantiations of GMRES and CG, which while not platform-specific, but likely represent the fundamental scaling properties of these solvers. Profiling a multi-chip instantiation of linear solvers would require system specific tailoring, which is beyond the scope of this study.
3) The current NeuroFEM implementation does not halt at convergence, so we overestimate the number of timesteps required as our upper bound energy and time estimations. This may be increasing the absolute time costs of the model by a factor of 10x, but this overestimation will vary with model size.

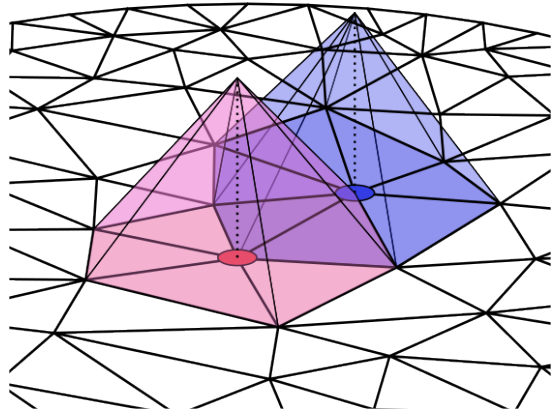

**Figure S1.**
Illustration of the piecewise linear elements used to construct the weak form of the PDEs.

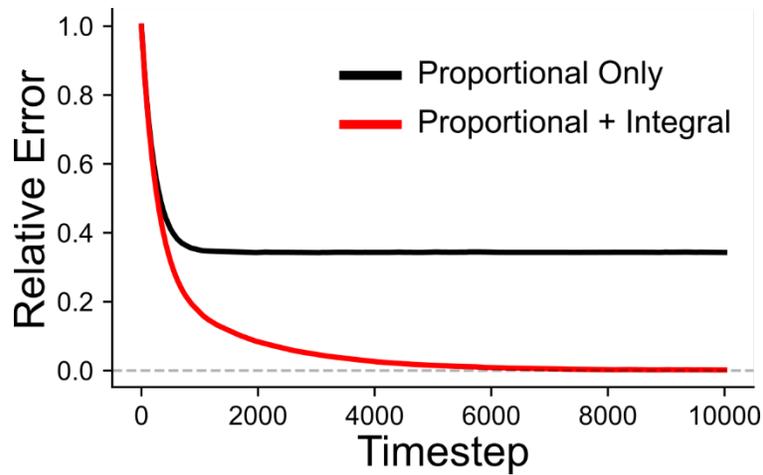

**Figure S2.**

The integral term is required for convergence. The black trace shows the relative error between the instantaneous NeuroFEM output and the conventional solution with only proportional terms [so the circuit is essentially equivalent to the model in (*9*)]. The circuit necessarily has a steady-state error. The red trace shows the same quantity but includes the integral term. This circuit converges to the solution.

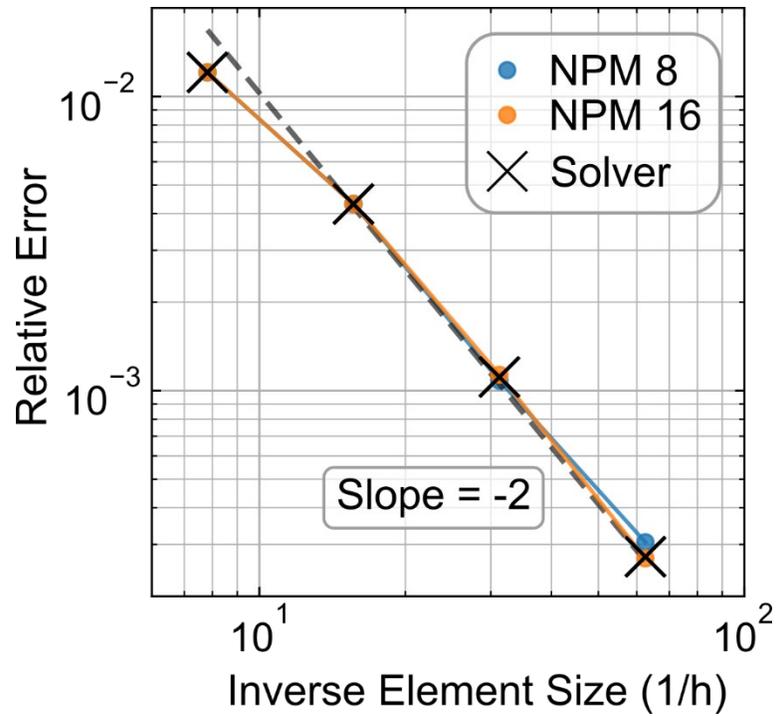

**Figure S3**
Convergence of NeuroFEM and the conventional solver to the analytic solution as a function of the element size. For piecewise linear (order 1) elements, finite element theory predicts a quadratic convergence rate, meaning the relative error should scale as the square of the element size. On the log-log plot shown here, both NeuroFEM and the conventional solver fall on a line with slope -2, agreeing with the theory. NPM: neurons per mesh node.

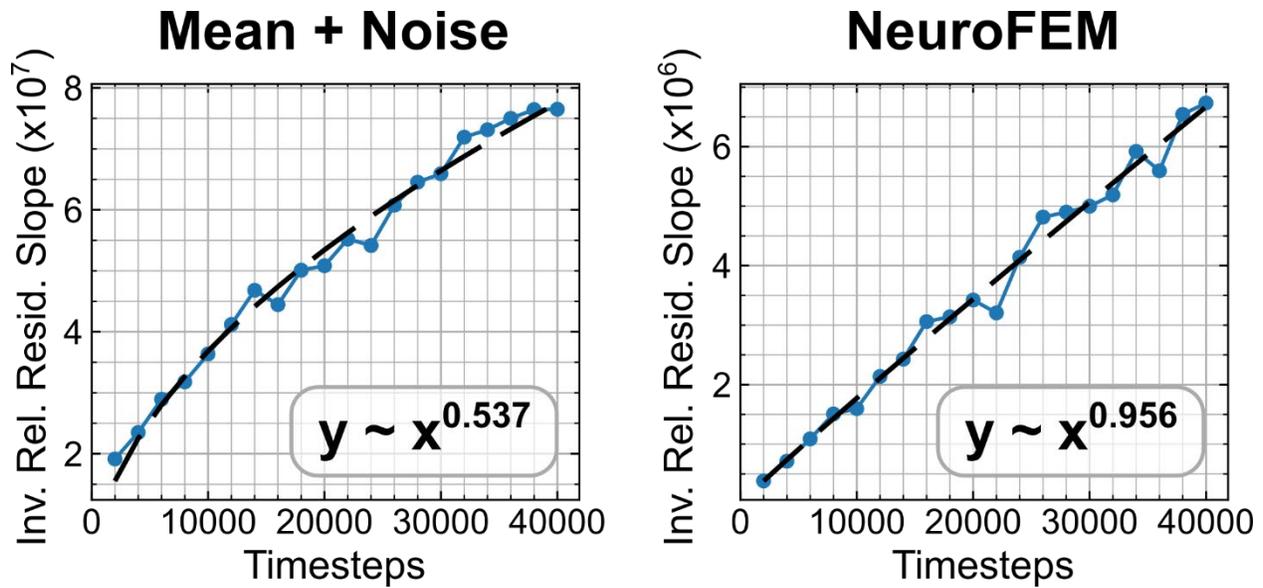

**Figure S4.**

NeuroFEM does not use a rate code to represent values. Left, Gaussian noise was added to the solution from a conventional solver and the inverse relative residual slope was computed for averages over different numbers of timesteps. As expected, this quantity improves proportional to the square root of the number of timesteps. Right, the fluctuating output from NeuroFEM improves linearly with respect to the number of timesteps in the average. Coordination between neurons in NeuroFEM requires that the timing of spikes matters to the representation, unlike a collection of independent rate-based readouts. Thus, the NeuroFEM readout is not simply a mean plus independent noise.

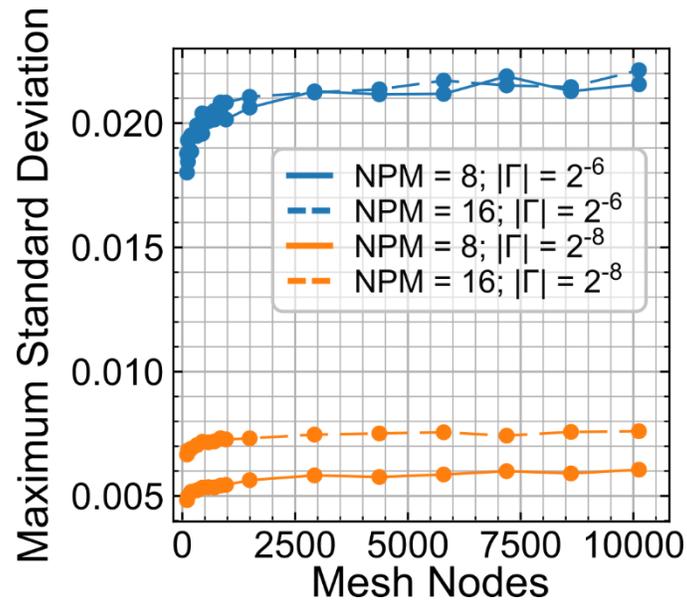

**Figure S5.**
Fluctuations in the readout do not grow with mesh size. Plotted is the maximum standard deviation of the readout over the last 10,000 timesteps across all mesh nodes (readout variables). This quantity stays approximately constant as the mesh size grows. Smaller $|\Gamma|$ yields smaller fluctuations.

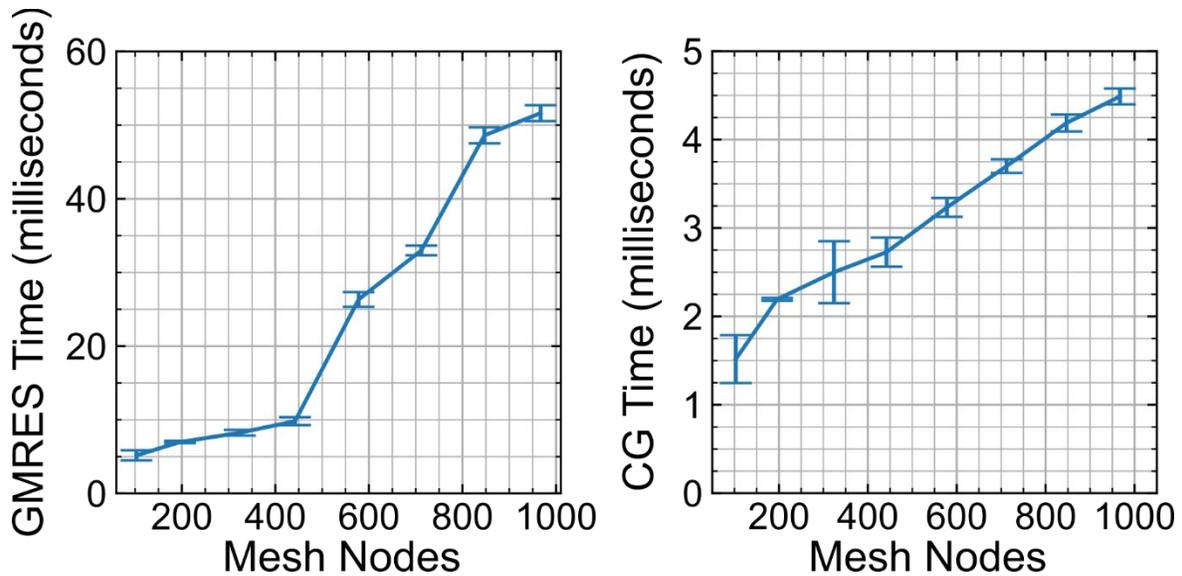

**Figure S6**
Time required for iterative solvers on CPU to solve the same FEM systems as in Figure 3. We profiled SciPy's generalized minimal residual algorithm (GMRES) (*35*) and Conjugate Gradient (CG) (*36*) solvers to solve these systems to a relative tolerance of $10^{-3}$. Error bars indicate standard deviation across 128 separate runs. Both algorithms scale similarly to our Loihi 2 results. GMRES is approximately 2x as fast as NeuroFEM on Loihi 2, and CG is approximately 20x as fast, though it must be noted that CG is specific to symmetric matrices, which is not a requirement of GMRES or our NeuroFEM method. Using an estimate of 100 Watts for the power draw of the conventional CPU, both algorithms require significantly more energy than NeuroFEM on Loihi 2 (~5 Joules for GMRES and ~0.45 Joules for CG as opposed to ~0.08 Joules for Loihi 2 for the largest system).

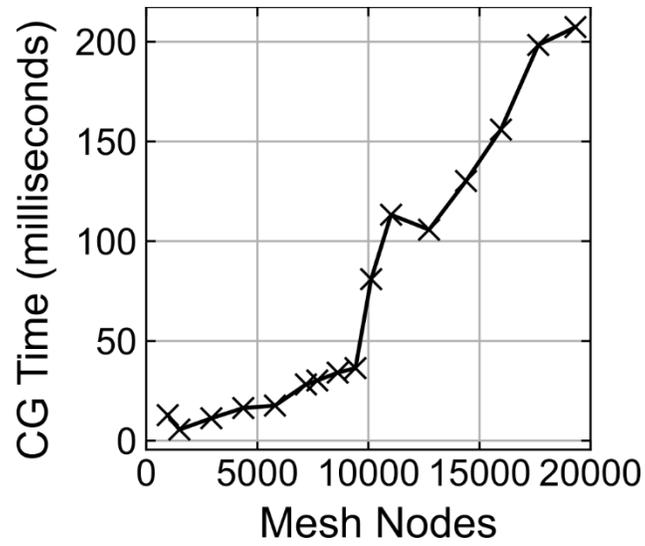

**Figure S7**
Time required for conjugate gradient (Scipy) on CPU to solve the same FEM linear systems as in the weak scaling study of Figure 4. CG was run until a relative tolerance of 1e-3.

**Table S1.**

Energy differences between transient and steady-state circuits on Loihi 2. This is the data plotted in Figure 3G.

| Mesh Resolution (Nodes) | Number of Neurons per Mesh Node | Mean Energy Difference (Joules) | Standard Deviation (Joules) |
|---|---|---|---|
| 103 | 8 | 0.0587 | 0.00301 |
| 103 | 16 | Did Not Converge | Did Not Converge |
| 196 | 8 | 0.0165 | 0.000895 |
| 196 | 16 | 0.0888 | 0.00329 |
| 324 | 8 | 0.0192 | 0.00174 |
| 324 | 16 | 0.0312 | 0.00187 |
| 442 | 8 | 0.0262 | 0.00213 |
| 442 | 16 | 0.0391 | 0.00184 |
| 578 | 8 | 0.0314 | 0.00173 |
| 578 | 16 | 0.0493 | 0.00284 |
| 712 | 8 | 0.0385 | 0.00191 |
| 712 | 16 | 0.0647 | 0.00203 |
| 847 | 8 | 0.0465 | 0.00173 |
| 847 | 16 | 0.0737 | 0.00165 |
| 967 | 8 | 0.0482 | 0.00239 |
| 967 | 16 | 0.0813 | 0.00477 |

**Table S2.**

Energy differences between transient and steady-state circuits on Loihi 2. This is the data plotted in Figure 3I.

| Mesh Resolution (Nodes) | Number of Neurons per Mesh Node | Mean Time Difference (Milliseconds) | Standard Deviation (Milliseconds) |
|---|---|---|---|
| 103 | 8 | 23.8 | 0.951 |
| 103 | 16 | Did Not Converge | Did Not Converge |
| 196 | 8 | 6.67 | 0.454 |
| 196 | 16 | 35.1 | 1.45 |
| 324 | 8 | 7.94 | 0.270 |
| 324 | 16 | 12.0 | 0.229 |
| 442 | 8 | 10.5 | 0.184 |
| 442 | 16 | 15.9 | 0.223 |
| 578 | 8 | 12.8 | 0.170 |
| 578 | 16 | 20.1 | 0.260 |
| 712 | 8 | 15.1 | 0.150 |
| 712 | 16 | 24.4 | 0.215 |
| 847 | 8 | 17.6 | 0.235 |
| 847 | 16 | 28.7 | 0.200 |
| 967 | 8 | 19.5 | 0.0664 |
| 967 | 16 | 32.2 | 0.161 |

**Movie S1.**

Animation of NeuroFEM solving the example problem from Figure 1. The top left panel shows a spike raster of the activity from 5 arbitrary nodes in the mesh. Red spikes contribute positive kernels to the readout and blue spikes contribute negative kernels. The bottom left panel shows the relative residual with respect to either $f_1$ (orange) or $f_2$ (blue). The right panel shows the instantaneous readout of the NeuroFEM circuit on the original mesh.